%% file: main.tex
\documentclass[sigconf, screen]{acmart}
\AtBeginDocument{%
  }

\setcopyright{acmlicensed}
\copyrightyear{2026}
\acmYear{2026}
\acmDOI{XXXXXXX.XXXXXXX}
\acmConference[Conference acronym 'XX]{ XX}{XX}{XXXX}

\acmISBN{978-1-4503-XXXX-X/2018/06}

\usepackage{multirow}
\usepackage{array}
\usepackage{booktabs}
\usepackage{etoolbox}
\usepackage{soul}      
\usepackage{xcolor}   
\usepackage{xspace}
\usepackage{relsize}
\usepackage{pifont} 
\usepackage{acronym}

\usepackage{subcaption}
\usepackage[table]{xcolor}
\colorlet{shadegray}{gray!10}
\definecolor{darkgreen}{rgb}{0,0.5,0}
\newcommand{\xmark}{\ding{55}}

\newcommand{\SysName}{ECG-Scan\xspace}
\usepackage{siunitx}
\makeatletter
\newcommand{\@toptitlebar}{
  \hrule height 4\p@
  \vskip 0.15in
  \vskip -\parskip%
}
\newcommand{\@bottomtitlebar}{
  \vskip 0.29in
  \vskip -\parskip
  \hrule height 1\p@
  \vskip 0.29in%
}
\makeatother

\newbool{showComments}
\setbool{showComments}{true}  

\ifbool{showComments}{

}

\begin{document}

\title{Learning ECG Image Representations via Dual Physiological-Aware Alignments}

\author{Hung Manh Pham}
\email{hm.pham.2023@phdcs.smu.edu.sg}

\affiliation{%
  \institution{University of Cambridge \\
  Singapore Management University}
  \country{}
}

\author{Jialu Tang}
\email{j.tang@tue.nl}
\affiliation{%
  \institution{Eindhoven University of Technology}
  \country{Netherlands}
}

\author{Aaqib Saeed}
\email{a.saeed@tue.nl}
\affiliation{%
  \institution{Eindhoven University of Technology}
    \country{Netherlands}
}

\author{Dong Ma}
\authornote{Equal Corresponding Author}
\email{dm878@cam.ac.uk}
\affiliation{%
  \institution{University of Cambridge}
  \country{United Kingdom}
}

\author{Bin Zhu}
\authornotemark[1]
\email{binzhu@smu.edu.sg}
\affiliation{%
  \institution{Singapore Management University}
    \country{Singapore}
}

\author{Zhou Pan}
\authornotemark[1]
\email{panzhou@smu.edu.sg}
\affiliation{%
  \institution{Singapore Management University}
    \country{Singapore}
}

\renewcommand{\shortauthors}{Pham et al.}

\input{sections/Abstract}

\ccsdesc[500]{Applied computing~Health informatics}
\ccsdesc[500]{Computing methodologies~Self-supervised learning}


\keywords{Cardiovascular Diagnostics, ECG-Image Foundation Models, ECG-Text Representation Learning}


\maketitle

\input{sections/Introduction}

\input{sections/Background}
\input{sections/Related-Work}
\input{sections/Methods}
\input{sections/Implementation}
\input{sections/Results}

\input{sections/Conclusion}





\bibliography{references}
\bibliographystyle{ACM-Reference-Format}

\input{sections/Appendix}

\end{document}

%% file: sections/Abstract.tex
\begin{abstract}
Electrocardiograms (ECGs) are among the most widely used diagnostic tools for cardiovascular diseases, and a large amount of ECG data worldwide appears only in \textit{image} form. However, most existing automated ECG analysis methods rely on access to raw \textit{signal} recordings, limiting their applicability in real-world and resource-constrained settings. In this paper, we present \SysName, a self-supervised framework for learning clinically generalized representations from ECG images through dual physiological-aware alignments: 1) Our approach optimizes image representation learning using multimodal contrastive alignment between image and gold-standard signal-text modalities. 2) We further integrate domain knowledge via soft-lead constraints, regularizing the reconstruction process and improving signal lead inter-consistency. Extensive experiments across multiple datasets and downstream tasks demonstrate that our image-based model achieves superior performance compared to existing image baselines and notably narrows the gap between ECG image and signal analysis. These results highlight the potential of self-supervised image modeling to unlock large-scale legacy ECG data and broaden access to automated cardiovascular diagnostics.
\end{abstract}

%% file: sections/Introduction.tex
\section{Introduction}

Cardiovascular diseases (CVDs) remain the leading cause of mortality worldwide, accounting for more than 20 million deaths annually, with approximately 80\% occurring in low- and middle-income countries~\cite{di2024heart}. Therefore, early detection, continuous monitoring, and accurate diagnosis of CVDs are critical to reducing mortality and improving patient outcomes. In this context, reliable and widely accessible diagnostic modalities are essential. 

\begin{figure*}[t]
    \centering
    \includegraphics[width=0.92\textwidth]{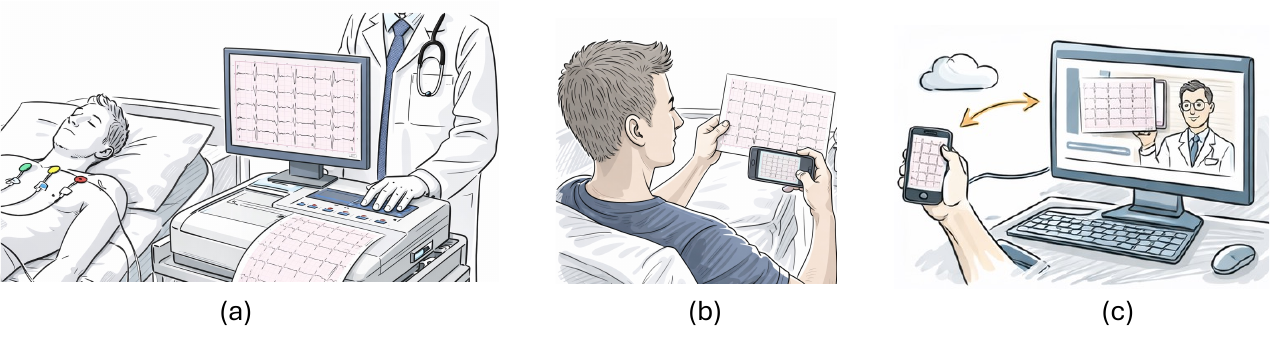}
    \vspace{-0.2cm}
    \caption{(a) Common ECG acquisition in a clinical environment, where an ECG machine is connected to a printer to get a paper-based document for patients. Optionally, a local scanner might be connected for digital archive (images, in PDF), depending on the ECG machine types and hardware systems. 
    (b) Patient capturing of ECG printouts using a smartphone camera (or a scanner), resulting in an image-based ECG for their own long-term archive. 
    (c) Possible remote review of ECG images during consultation service, where expert clinicians interpret in-depth cardiac patterns from shared ECG images.}
    \label{fig:scnarios}
\end{figure*}

Among available modalities, the electrocardiogram (ECG) is widely regarded as the canonical standard for non-invasive cardiac diagnosis. Accordingly, since the invention of the first practical ECG device by Willem Einthoven in 1895~\cite{barold2003willem}, ECG technology has undergone significant evolution. The introduction of portable ECG devices in the early twentieth century and the widespread adoption of paper-based waveform recordings by the mid-century enabled ECG examination to become routine in clinical practice~\cite{reyna2024digitization}. To date, despite substantial progress in digital ECG systems and algorithmic interpretation, ECGs are still predominantly used and stored as printed or image-based records in many real-world settings, with billions of paper ECG samples around the world~\cite{opendigitizer}, particularly in resource-limited regions and across the Global South~\cite{tison2019automated, reyna2024digitization, sarahretrospective, ecgkit}. Therefore, the ability to interpret ECG \textit{images} is essential for unlocking these data and improving equitable access to cardiac care. 

Despite this widely recognized reality, existing ECG analysis methods still largely assume direct access to raw 12-lead \textit{signal} recordings. Among them, the recent methods showed strong performance for ECG signal representation learning using ECG self-supervised learning (SSL), particularly when augmented with multimodal information such as clinical text~\cite{liu2024zero, anychat, melp, dbeta, htune}. These advances further raise a natural question: \textit{Can SSL be extended to ECG image-text learning to produce generalized image representations that approach the fidelity and utility of signal-based representations?} Several works have explored image-based ECG analysis and applications such as supervised classification~\cite{gliner2025clinically}, image-to-signal conversion pipelines~\cite{nnunet, opendigitizer}, or multimodal vision-language modeling~\cite{pulse, gem} for textual cardiac interpretation. 

While these approaches demonstrate promise, they either rely on generic visual encoders or small-scale supervised digitization pipelines that require handcrafting processing steps and remain limited in capturing the full temporal and physiological richness of gold-standard ECG signals. Therefore, existing image-based modeling often shows suboptimal performance, compared to signal-text works, which are well-demonstrated for strong generalization across tasks and datasets~\cite{liu2024zero}. From here, we also observe one key challenge that ECG images encode temporal cardiac dynamics only implicitly through spatial layouts and often provide incomplete temporal coverage per lead (e.g., 2.5 seconds long). \textit{As a result, learning robust and transferable representations directly from ECG images is considerably more challenging than learning from 12-lead 10-second signals as gold standards.}

In this paper, we aim to address the gap by \textit{unifying image, signal, and text modalities within a self-supervised framework}. Rather than relying on direct supervision, general-image encoders, or brittle digitization pipelines, our models directly learn generalized ECG image representations for efficient downstream tasks by aligning them with underlying strong multimodal signal-text representations and enforcing physiological consistency through domain-informed constraints. Starting from a published large-scale ECG signal-text dataset~\cite{gow2023mimic}, we synthetically generate diverse ECG images that emulate real-world printouts, enabling scalable pretraining without manual annotation. As we have three modalities, we perform multimodal physiological alignment learning by jointly aligning image, signal, and text representations in a shared latent space and reconstructing full 12-lead ECG signals from images. 

We then propose dual physiological-aware alignments with two key components: 1) Align these three modalities with the Gramian-based contrastive learning method while keeping boosted contrastive image-text alignment, which preserves semantic interpretability at inference time when signals are unavailable; 2) Signal reconstruction anchors image representations physiologically consistent with temporal and morphological structure. We introduce soft lead-consistency regularization that incorporates established physiological constraints from Einthoven’s law~\cite{barold2003willem} and Goldberger’s lead relationship~\cite{goldberger1942avl} equations into the learning objective. This domain-informed regularization improves the physiological plausibility of reconstructed signals and stabilizes representation learning. Together, these designs enable robust ECG image representations that approach the fidelity of signal-based models while remaining applicable in image-only settings.

Our contributions can be summarized as follows:
\begin{itemize}
    \item We introduce the first ECG image self-supervised framework that learns visual-based ECG representations, approaching the diagnostic performance of state-of-the-art 12-lead signal-based foundation models. 
    \item We propose a dual physiological-aware alignment strategy that enforces consistency in both latent space and time-series space, leveraging an enhanced Gramian-based contrastive alignment and Einthoven-Goldberger soft lead consistency alignment, respectively.
    \item We conduct a comprehensive evaluation across multiple datasets, demonstrating the effectiveness of the learned ECG image representations. We will release code and checkpoints to support reproducibility and future research.
\end{itemize}

%% file: sections/Background.tex
\section{Background}

The 12-lead ECG signals can capture cardiac electrical activity across multiple anatomical planes and have served as the clinical gold standard for cardiovascular diagnosis for decades. It comprises six limb leads (I, II, III, aVR, aVL, aVF) and six precordial (chest) leads (V1–V6), with the chest leads placed sequentially across the thorax to capture transverse-plane cardiac activity (see our Figure~\ref{fig:leads}). Together, electrodes positioned on the limbs and chest provide spatially diverse projections of the cardiac electrical vector, enabling comprehensive assessment of rhythm, conduction, and myocardial abnormalities.

While these digital signal-based ECG systems are increasingly adopted, ECG interpretation in routine clinical practice remains strongly tied to printed or image-based records. In many real-world and retrospective scenarios, raw digital signals are unavailable, rendering existing models inapplicable or impractical for deployment~\cite{opendigitizer}. \textit{This reliance on image-based ECGs is particularly pronounced in resource-constrained, rural, and remote healthcare settings, where ECGs are frequently archived as paper printouts or scanned images without accompanying signal repositories, while the local clinicians might be less experienced in ECG expertise}. We provide three practical scenarios that support our points on the usage of ECG images in Figure~\ref{fig:scnarios}.

Furthermore, it is widely known that \textit{cardiologists have reliably interpreted ECGs in visual form for decades, demonstrating that diagnostically meaningful cardiac information is preserved in the image domain itself}. In fact, vast collections of historical ECGs accumulated over time existed exclusively in image form, and cardiologists still routinely interpret visually, such as observing rhythm regularity and morphology patterns. From a systems and accessibility standpoint, image-based ECG data are also substantially easier to acquire, store, and share using commodity devices such as scanners or smartphone cameras, without reliance on vendor-specific ECG management systems. In contrast, proprietary “walled-garden” ECG infrastructures impose significant barriers to signal data access, interoperability, and large-scale analysis~\cite{reyna2024digitization}. 

This paper seeks to bridge this gap by enabling models to acquire representations that approach this human interpretability, learning directly from ECG images rather than requiring explicit signal reconstruction or proprietary data access. In the section below, we present more related works relevant to our study.


%% file: sections/Related-Work.tex
\section{Related Work}
\subsection{ECG Signal Representation Learning}

Recent years have seen a strong research focus on ECG representation learning based on raw time-series signals~\cite{hu2023spatiotemporal, nguyen2025tolerantecg, anychat, esi, heartlang, ecgfm, ecgfounder}. Among them, several large-scale self-supervised and multimodal frameworks have demonstrated effective and practical performance across a broad range of downstream tasks, including arrhythmia classification, zero-shot inference, clinical question answering, and report generation~\cite{liu2024zero, dbeta, melp, ecgqa, qheart}, collectively establishing robust signal-domain foundations for ECG signal analysis. However, despite growing adoption of digital ECG systems, routine clinical practice and large retrospective archives remain heavily reliant on printed or image-based ECG records, especially in resource-constrained settings where raw signals are unavailable.

\subsection{ECG Image Modeling}

To address the limitations of signal-only methods, a growing body of work explores ECG analysis directly from images~\cite{sangha2022automated, gliner2025clinically, pulse, gem}. Early efforts, such as ~\cite{sangha2022automated, gliner2025clinically}, focus on supervised learning for ECG image classification, demonstrating that clinically relevant information can be extracted from printed forms. Following that, more recent and robust methods are multimodal large language modeling (MLLM) approaches~\cite{pulse, gem} that extend the paradigm by jointly modeling ECG images and text for tasks such as report generation and visual question answering. Despite encouraging results, most existing image-based methods rely on generic vision encoders or frozen image backbones~\cite{pulse, gem}, originally well-developed for natural images (e.g., the pretrained CLIP image encoder~\cite{clip} from LLaVA~\cite{liu2023visual}), rather than the ECG images. Such encoders are then less effectively aligned with the structural properties of ECG images, which encode dense temporal waveforms arranged spatially rather than semantic objects. Consequently, these models often exhibit limited robustness in representation learning for various downstream tasks compared to signal foundation models.

\subsection{Image-to-Signal Conversions}

Another line of work seeks to bridge the modality gap by converting ECG images back into signal representations prior to analysis~\cite{baydoun2019high, wu2022fully, nnunet, opendigitizer}. As a natural approach, there are classical ways that rely on a series of image processing (DIP) steps, such as template \& layout matching, edge \& point detection, or grid removal, and finally, lead detection \& extraction pipelines, which are highly sensitive to noise, grid variability, and printing artifacts. Moving forward, more recent hybrid methods combine both those DIP techniques and deep segmentation models~\cite{nnunet, opendigitizer}, such as U-Net-style architectures, to improve robustness across more input formats. However, while image-to-signal conversion enables reuse of existing signal-based models, current methods remain limited by small-scale paired image-signal supervision and typically learn to reconstruct only shortened signals present in ECG images. As a result, they struggle with waveform diversity and generalization, ultimately constraining downstream performance and adaptation capability even when coupled with strong signal-based models.
 

%% file: sections/Methods.tex
\section{Methods}
\label{methods}

\begin{figure*}[t]
    \centering
    \includegraphics[width=\textwidth]{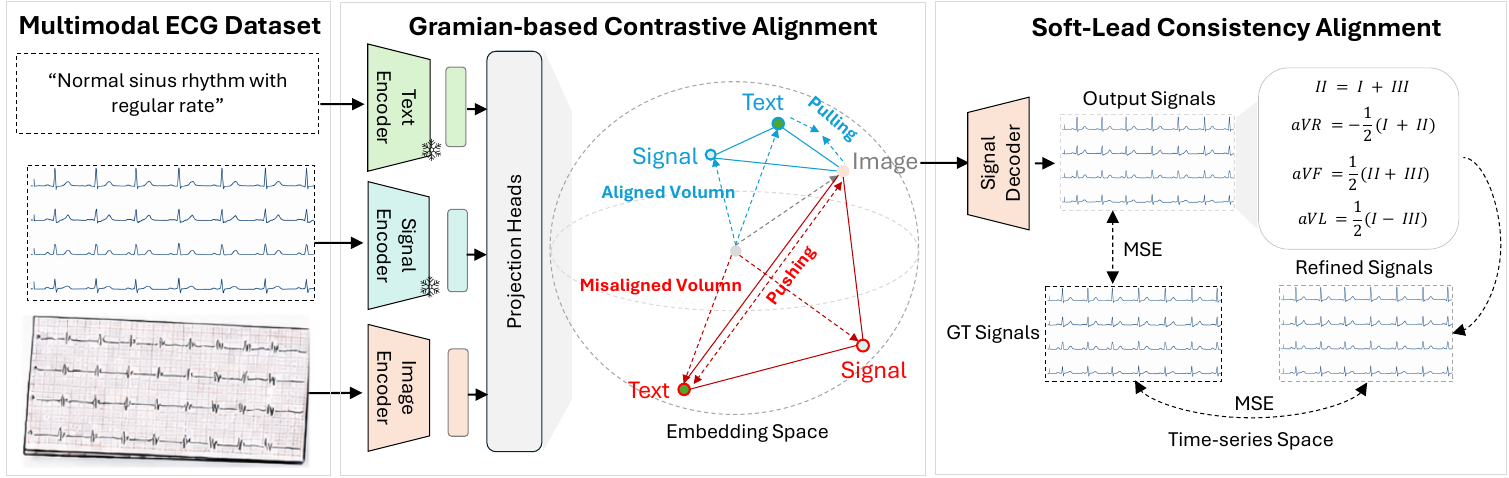}
    \caption{Illustration of \SysName. We present a multimodal framework that aligns ECG images, signals, and clinical texts through dual physiological-aware alignment strategy.}
    \label{fig:overview}
\end{figure*}

In this section, we present \SysName, a self-supervised framework for learning ECG image representations that leverages ECG images, signals, and clinical text during pretraining, illustrated in Figure~\ref{fig:overview}. \SysName consists of three model components: 1) an ECG image encoder that extracts visual features from ECG images, 2) frozen well-trained signal-text encoders that provide teacher representations, and 3) a signal decoder that reconstructs 12-lead ECG signals from image representations. In general, \SysName uses ECG signals and text reports as privileged supervision during pretraining to guide the image encoder, with a dual physiological-aware alignment strategy that enforces consistency between ECG images, signals, and text in both (i) the latent representation space and (ii) the reconstructed time-series space. This design allows the image encoder to capture fine-grained cardiac information from visual ECG patterns as well as inherit clinically discriminative representations from signal-text-based foundation models. In the below sections, we provide further detailed descriptions of our framework.

\subsection{Problem Formulation}

Let $\mathcal{X}_{\text{img}} \in \mathbb{R}^{H \times W \times 3}$ denote an ECG image, $\mathcal{X}_{\text{sig}} \in \mathbb{R}^{C \times T}$ denote the corresponding 12-lead ECG signal with $C=12$ leads and $T$ time steps, and $\mathcal{X}_{\text{txt}}$ denote the associated clinical text report. Our objective is to learn an ECG image encoder $f_{\theta}$ that produces high-quality ECG representations from images alone, by aligning image representations with their paired ECG signals ($z_{\text{sig}}$) and text descriptions ($z_{\text{txt}}$) during pretraining. Here, the key motivation is that ECG signals are the gold standard for cardiology analysis, while ECG text reports capture high-level diagnostic semantics routinely used in clinical decision-making. We aim to learn an image encoder that outputs signal-level cardiac morphology and diagnostic semantics within visual representations ($z_{\text{img}}$) when ECG signals are unavailable at inference time.

\subsection{Gramian-based Contrastive Alignment}

We present aligning modality representations through a combination of two approaches: pairwise image-text contrastive learning and extension with a Gramian-based loss that enforces three-way geometric consistency across image, text, and signal modalities.  When ECG images are used as the main input, on the one hand, image-text contrastive learning methods will only find it challenging to yield representations that capture physiologically meaningful signal structure, as large portions of the image contain redundant visual elements unrelated to the underlying cardiac waveform. Therefore, signal and text representations with highly related cardiac information can support guiding the training process. On the other hand, while the Gramian-based method~\cite{gramian} focuses on enforcing higher-order geometric consistency across multiple modalities, it is not designed to provide strong discriminative supervision between samples (e.g., images vs. texts for zero-shot learning). When applied alone, such objectives may lead to insufficient separation between clinically distinct ECG patterns that share similar physiological structure. Therefore, we use Gramian alignment as a physiological regularizer, while retaining image-text contrastive learning to explicitly enable discriminative power during signal-free inference.

\textit{Image-Text Contrastive Loss.} Firstly, we use standard image-text contrastive learning to strongly align ECG images with clinical text. Following previous works on leveraging the strength of clinical texts~\cite{liu2024zero, melp, dbeta} to support ECG signal learning, we align ECG image and text representations: Given a batch of image-text pairs with projected representations $\mathbf{z}_{\text{img}}^{\text{ctr}} = \text{Proj}_{\text{ctr}}(\mathbf{z}_{\text{img}})$ and $\mathbf{z}_{\text{txt}}$, we compute contrastive loss following~\cite{clip}:

\begin{equation}
\mathcal{L}_{\text{ctr}}
=
\frac{1}{2}
\Big(
\mathrm{CE}(\tau\, Z_{\text{img}} Z_{\text{txt}}^{\top})
+
\mathrm{CE}(\tau\, Z_{\text{txt}} Z_{\text{img}}^{\top})
\Big),
\end{equation}

where CE denotes cross-entropy with label smoothing (0.1) and $\tau$ is a learnable temperature parameter.

\textit{Gramian Three-Way Alignment.} We encourage the image representation to also be consistent with the signal representation from a well-trained signal encoder. Specifically, signal representations encode rich temporal information that directly reflects cardiac physiology from raw signal data. The Gramian-based alignment leverages this property by distilling higher-order relational structure from signal embeddings into the image encoder, acting as a physiological regularizer. We achieve this through a Gramian-based volume loss that measures the geometric alignment of all three modalities simultaneously. Given normalized embeddings $\tilde{\mathbf{z}}_{\text{img}}$, $\tilde{\mathbf{z}}_{\text{txt}}$, and $\tilde{\mathbf{z}}_{\text{sig}}$, we compute the volume of the parallelepiped spanned by these vectors using the Gram determinant:
\begin{equation}
    V(\tilde{\mathbf{z}}_{\text{img}}, \tilde{\mathbf{z}}_{\text{txt}}, \tilde{\mathbf{z}}_{\text{sig}}) = \sqrt{\left| \det(\mathbf{G}) \right|},
\end{equation}
where $\mathbf{G}$ is the Gram matrix:
\begin{equation}
    \mathbf{G} = \begin{pmatrix}
        \langle \tilde{\mathbf{z}}_{\text{img}}, \tilde{\mathbf{z}}_{\text{img}} \rangle & \langle \tilde{\mathbf{z}}_{\text{img}}, \tilde{\mathbf{z}}_{\text{txt}} \rangle & \langle \tilde{\mathbf{z}}_{\text{img}}, \tilde{\mathbf{z}}_{\text{sig}} \rangle \\
        \langle \tilde{\mathbf{z}}_{\text{txt}}, \tilde{\mathbf{z}}_{\text{img}} \rangle & \langle \tilde{\mathbf{z}}_{\text{txt}}, \tilde{\mathbf{z}}_{\text{txt}} \rangle & \langle \tilde{\mathbf{z}}_{\text{txt}}, \tilde{\mathbf{z}}_{\text{sig}} \rangle \\
        \langle \tilde{\mathbf{z}}_{\text{sig}}, \tilde{\mathbf{z}}_{\text{img}} \rangle & \langle \tilde{\mathbf{z}}_{\text{sig}}, \tilde{\mathbf{z}}_{\text{txt}} \rangle & \langle \tilde{\mathbf{z}}_{\text{sig}}, \tilde{\mathbf{z}}_{\text{sig}} \rangle
    \end{pmatrix}.
\end{equation}

Intuitively, the volume approaches a smaller value when the three vectors are well-aligned (i.e., lie in a low-dimensional subspace), indicating that the image representation captures information consistent with both the textual and signal-derived features. Specifically, the loss is computed using bidirectional cross-entropy over in-batch negatives:
\begin{equation}
    \mathcal{L}_{\text{gram}} = \frac{1}{2} \left( \text{CE}(-V \tau) + \text{CE}(-V^\top \tau \right).
\end{equation}

\subsection{Soft-Lead Consistency Alignment}

Beyond multimodal alignment, our framework also enforces physiological plausibility at the signal time-series level. This component leverages well-established ECG limb-lead relationships to regularize signal reconstruction, ensuring that reconstructed waveforms not only match the ground truth numerically but also preserve clinically meaningful inter-lead structure (see our Figure~\ref{fig:overview}). First, a standard mean squared error (MSE) loss is used to measure the fidelity of the decoded signal against the ground-truth 10-second 12-lead ECG. This reconstruction objective aims to help learning fine-grained electrophysiological structure rather than superficial visual patterns, encouraging the image encoder to capture foundational temporal morphology and waveform characteristics that are well-suited in clinical ECG settings:

\begin{equation}
    \mathcal{L}_{\text{mse}} = \frac{1}{C \cdot T} \sum_{c=1}^{C} \sum_{t=1}^{T} \left( \hat{\mathcal{X}}_{\text{sig}}^{(c,t)} - \mathcal{X}_{\text{sig}}^{(c,t)} \right)^2.
\end{equation} 

\begin{figure}[t]
    \centering
    \includegraphics[width=0.45\textwidth]{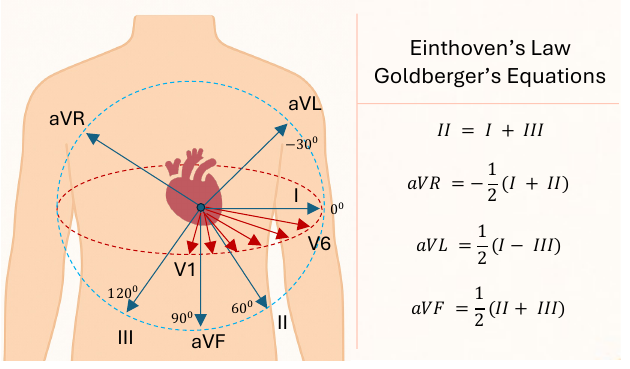}
    \caption{Illustration of Einthoven’s Law and Goldberger’s equations for limb leads of ECG signals.}
    \label{fig:leads}
\end{figure}

Next, while reconstruction loss enforces overall signal fidelity, it does not explicitly encode known physiological relationships among ECG leads. Specifically, in standard 12-lead ECGs, limb leads obey well-established electrophysiological constraints, including Einthoven’s law ~\cite{barold2003willem} and Goldberger’s equations~\cite{goldberger1942avl}, as shown in Figure~\ref{fig:leads}. In practice, however, these relationships are not satisfied exactly, as ECG signals and images may contain noise, distortions, or incomplete information arising from acquisition artifacts. Consequently, enforcing these constraints as hard rules may be overly restrictive and potentially destabilize training. To address this, we incorporate physiological knowledge through soft-lead consistency alignments. Rather than enforcing strict equality, soft constraints regularize reconstructed signals toward physiologically plausible configurations while allowing flexibility to accommodate real-world variability. This design improves robustness and stability during training and encourages reconstructions that are both realistic and physiologically consistent.

Based on the relationships, we define refined signals by projecting the reconstructed limb leads onto the corresponding constraint manifold. For example, for leads I, II, and III, the refined signals are computed as:

\begin{equation}\scriptsize
\text{I}_{\text{ref}}=\tfrac13(2\hat{\text{I}}+\hat{\text{II}}-\hat{\text{III}}),\;
\text{III}_{\text{ref}}=\tfrac13(-\hat{\text{I}}+\hat{\text{II}}+2\hat{\text{III}}),\;
\text{II}_{\text{ref}}=\text{I}_{\text{ref}}+\text{III}_{\text{ref}}.
\end{equation}

We then penalize deviations between ground truth signals and those physiologically derived refined signals using the following consistency loss:
\begin{equation}\small
    \mathcal{L}_{\text{rule}} = w_E \cdot \ell(\mathbf{X}_{\text{I,II,III; ref}}, \mathbf{X}_{\text{I, II, III}}) + w_G \cdot \ell(\mathbf{X}_{\text{aV; ref}}, \mathbf{X}_{\text{aV}}),
\end{equation}
where $\ell(\cdot, \cdot)$ denotes the MSE loss, and $w_E = w_G = 0.5$ balance the contributions from Einthoven and Goldberger constraints.

\textbf{Total Training Objective.} Finally, our overall training objective integrates discriminative multimodal alignment with knowledge-aware generative reconstruction:
\begin{equation}
    \mathcal{L} = \alpha \cdot \mathcal{L}_{\text{ctr}} + \theta \cdot \mathcal{L}_{\text{gram}} + \beta \cdot \mathcal{L}_{\text{recon}}, 
\end{equation}
where
\begin{equation}
    \mathcal{L}_{\text{recon}} = \mathcal{L}_{\text{mse}} + w_{\text{rule}} \cdot \mathcal{L}_{\text{rule}}.
\end{equation}

The hyperparameters $\alpha$, $\beta$, $\theta$, and $w_{\text{rule}}$ control the relative importance of contrastive alignment, reconstruction fidelity, and latent physiological regularization, respectively.

%% file: sections/Implementation.tex
\section{Datasets and Experimental Setup}

\subsection{Datasets}

\textbf{Pretraining Dataset.}
We pretrain our model on the MIMIC-IV-ECG dataset~\cite{gow2023mimic}, a large-scale clinical corpus containing paired 12-lead ECG signals (10 seconds at 500 Hz) and free-text diagnostic reports. We largely follow the data preprocessing steps of recent signal-based work~\cite{liu2024zero}, resulting in 789,511 signal-text pair samples, while extending them to include ECG images correspondingly. 

\textbf{Downstream Datasets.}
To evaluate the pretrained image encoder, we also follow recent ECG benchmarking protocols \cite{liu2024zero, dbeta, melp} and consider four widely used public datasets: PTB-XL~\citep{wagner2020ptb}, CSN~\citep{zheng2022large}, CPSC2018~\citep{liu2018open}, and CODE-test~\citep{ribeiro_automatic_2020}, which contain different ECG signals and numerous cardiac conditions as labels to evaluate. We preprocess and ensure the signals have the same length (10 seconds), lead orders, and a sampling rate of 500 Hz. It is also worth noting that the PTB-XL dataset itself has different types of labels: super-class, sub-class, form, and rhythm labels as four independent sub-datasets. After that, these processed signal datasets would be pre-generated for building corresponding downstream image datasets, while the labels are kept the same. 

Additionally, we provide Table~\ref{tab:snr} to support the perspective by reporting the mean signal-to-noise ratio (SNR) between limb leads reconstructed using Einthoven’s and Goldberger’s formulas and the corresponding recorded leads across three commonly used downstream ECG datasets (PTB-XL, CPSC2018, and CSN). The consistently high SNR values (typically exceeding 40-50 dB), which indicate that these physiological relationships are strongly preserved in real-world ECG recordings despite noise, acquisition artifacts, and dataset heterogeneity. This empirical observation justifies our use of soft lead consistency constraints during pretraining.

\begin{table}[t]
    \centering
    \setlength\tabcolsep{12pt}
    \caption{Mean SNR values across downstream datasets. Here, SNR was computed by comparing limb leads (III, aVR, aVL, and aVF) calculated from Leads I and II using Einthoven’s and Goldberger’s formulas with the actual recorded leads.}
    \label{tab:snr}
    \begin{tabular}{lccc}
        \toprule
        \textbf{Lead} & \textbf{PTB-XL} & \textbf{CPSC} & \textbf{CSN} \\
        \midrule
        Lead III & 50.70 & 89.62 & 46.41 \\
        Lead aVR & 48.65 & 45.61 & 36.54 \\
        Lead aVL & 47.38 & 41.89 & 33.55 \\
        Lead aVF & 47.13 & 43.95 & 35.70 \\
        \bottomrule
    \end{tabular}
    \vspace{-0.2pt}
\end{table}

\subsection{Experimental Setup}
\textbf{Pretraining.}
During pretraining, ECG signals from the MIMIC-IV-ECG dataset are dynamically converted into online augmented ECG images within each training step, simulating a diverse range of real-world ECG printouts with varying layouts, resolutions, and noise patterns. This is achieved by using a popular ECG plot toolkit~\cite{ecgkit}. More details can be found in our supplementary documents.

By default, we use D-BETA~\cite{dbeta} for the signal encoder and Bio-Med-CPT~\cite{jin2023medcpt} for the text encoder, which are frozen throughout pretraining, while the CLIP~\cite{clip} image encoder is used as the ECG image encoder and is adapted using low-rank adaptation (LoRA) with rank $r=16$ and scaling factor $\alpha=32$. For the signal decoder, we employ a Transformer-based encoder architecture with $L=12$ layers, hidden dimension $d=768$, and $H=12$ attention heads (more details are presented in our supplementary document). 

\SysName is trained on a single NVIDIA H200 GPU with a batch size of 80 and a learning rate of $5\times10^{-4}$, using the AdamW optimizer with a cosine learning rate scheduler and a 10\% warmup. In our experiments, we empirically chose $\alpha = 0.1$, $\beta = 1.0$, $\theta = 0.05$, and $w_{\text{rule}} = 0.1$ to balance component losses. Finally, the training proceeds for approximately 50,000 steps, and the checkpoint with the lowest validation loss is selected for downstream evaluation.

\textbf{Downstream Tasks.} We evaluate \SysName under two common complementary ECG downstream settings that assess representation quality: 1) Linear Probing Classification: We adopt a standard linear probing protocol in which the pretrained image encoder is frozen and a linear classifier is trained on top. Following common established evaluation pipelines~\cite{liu2024zero, melp, dbeta}, performance is reported using AUC (in \%) under different training sizes (1\%, 10\%, and 100\%) on PTB-XL, CSN, and CPSC2018. As different recent methods may have heterogeneous fine-tuning configurations~\cite{esi, heartlang,ecgfm, anychat,ecgfounder,melp,dbeta}, we re-implement their official model and use pretrained checkpoints whenever available for this task, as the first presented benchmark from MERL~\cite{liu2024zero}; 2) Zero-Shot Classification: Beyond supervised evaluation, we also assess zero-shot classification (using AUC in \%) on PTB-XL, CSN, CPSC2018, and CODE-test~\cite{liu2024zero, melp, dbeta}. In this setting, ECG representations are matched against text embeddings derived from possible context-enhanced diagnostic categories~\cite{liu2024zero, melp, dbeta}. 

\textbf{Baselines.} We compare \SysName against three key types of baselines: 1) \textit{Signal-Based Baselines:} Operating directly on 10s ECG signals and serving as \textbf{upper bounds for downstream performance}~\cite{liu2024zero, melp, dbeta, anychat, esi, heartlang, ecgfm, ecgfounder}; 2) \textit{Image-to-Signal Baselines:} Converting ECG images into signals before using the state-of-the-art masked signal-based foundation model (i.e., ~\cite{dbeta}). Here, we consider both the traditional digital image processing digitization (DIP) and recent methods with supervised U-Net-style segmentation~\cite{nnunet, opendigitizer} for the conversion, including nnUnet~\cite{nnunet} that won the George B. Moody PhysioNet Challenges~\cite{reyna2024digitization}; 3) \textit{Image-Only Baselines:} Learning representations from ECG images without explicitly reconstructing the signal. We include general-purpose and medical image encoders, such as CLIP~\cite{clip} and MedSigLIP~\cite{medsiglip}.


%% file: sections/Results.tex
\section{Results}

\input{tables/linear_probing}

\input{tables/zero-shot}

\input{tables/domain_shift}

\input{tables/ablations}

\subsection{Linear Probing Evaluation}

Table~\ref{tab:linear} reports linear probing performance in PTB-XL, CPSC2018, and CSN datasets under varying proportions of labeled data for downstream fine-tuning. On average, \SysName consistently outperforms generic image baselines and image-to-signal pipelines across all datasets and supervision regimes, while substantially narrowing the performance gap to strong signal-based foundation models. In particular, \SysName achieves approximately a 3\% absolute improvement over nnU-Net and Open-Digitizer + D-BETA in the 10\% and 100\% labeled data settings, highlighting the benefit of learning diagnostically meaningful representations directly from ECG images. Meanwhile, performance differences in the 1\% regime are relatively small, where image-based and digitization-based approaches exhibit comparable behavior. We attribute this to the fact that D-BETA benefits from broad pretraining on masked ECG signals, which provides stronger inductive bias when downstream supervision is extremely limited, but becomes less advantageous as more labeled data (e.g., 10\%, 100\%) are available for adaptation.

From Table~\ref{tab:linear}, we further observe that \SysName compares favorably against a wide range of signal-based foundation models (using full 10-second inputs), despite operating purely on ECG images (2.5 seconds, except lead II). For example, \SysName consistently outperforms several strong signal foundations such as MERL, which achieves averaged AUCs of 66.0\%, 81.7\%, and 86.7\% under the 1\%, 10\%, and 100\% supervision regimes, respectively, whereas \SysName attains 71.9\%, 84.5\%, and 90.4\% under the same settings. Moreover, \SysName substantially narrows the performance gap to state-of-the-art signal-based models, including ECGFounder, MELP, and D-BETA, which are trained directly on full-resolution ECG signals and represent current upper bounds in linear probing performance. This trend indicates that our approach is able to output relevant information from ECG images alone, yielding representations that are increasingly comparable with leading signal-based foundations as downstream supervision increases.

\subsection{Zero-shot Evaluation}

For zero-shot classification, we first evaluate diagnostic classification performance across PTB-XL, CPSC2018, and CSN. As shown in Table~\ref{tab:zerozhot1}, \SysName achieves an average AUC of 75.8\%, outperforming all image-to-signal baselines. Specifically, \SysName substantially improves over classical DIP + D-BETA (63.2) and nnU-Net + D-BETA (66.5), demonstrating the same observation in the linear probing experiments. While signal-based multimodal foundation models such as MELP remain upper bounds, \SysName closely reaches their top despite relying solely on ECG images at inference time, especially slightly over the MERL (75.3\%).

Similarly, we evaluate zero-shot ECG diagnosis by comparing \SysName against human experts, signal-based models, and image-based baselines, as summarized in Table~\ref{tab:zeroshot2}. \SysName achieves an AUC of 94.78\%, surpassing different human readers, including cardiology residents (92.07\%), emergency residents (90.52\%), and medical students (93.61\%). Here, medical students outperform residents, likely reflecting their more recent and focused training, similarly as reported in prior work~\cite{ribeiro_automatic_2020}. Moreover, \SysName achieves strong performance relative to signal foundation models such as MERL (only 85.14\%). Notably, in this experiment, \SysName clearly outperforms prior image-based approaches that use DIP digitization or U-Net-based backbones, even with the help of the strong signal encoders (e.g., DIP + D-BETA, nnUNet + D-BETA), further confirming our model's ability to yield diagnostically useful representations directly from ECG images, both closely comparable with human experts and other signal-based models.

Finally, we evaluate zero-shot performance against various signal foundation model baselines under domain shift settings, as shown in Table~\ref{tab:domainshift}. Specifically, we follow~\cite{liu2024zero} in their same experiments to compare our zero-shot performance to those models under their linear probing fine-tuning using 100\% of source data to test on the target ECG data (only data with mappable classes are used to evaluate). From the table, we can observe that \SysName achieves the average AUC of 80\%, interestingly surpassing all of the signal baselines in this experiment (e.g., surpassing ST-MEM~\cite{hu2023spatiotemporal} nearly 8\% and slightly over MELP~\cite{melp} at 79.6\%). We attribute this to the inherently strong transferable ability in the CLIP image encoder in our multimodal alignments, as well as the effectiveness of the diverse stochastic image augmentations applied during pre-training. 

\subsection{Ablation Studies}

In this section, we conduct ablations to quantify the impact of each core component, assess sensitivity to the choice of image/text/signal encoders, and provide t-SNE visualizations to qualitatively examine the learned embeddings. The evaluation reports linear probing in 1\% case and zero-shot classification, averaged over the six datasets. 

\textbf{Impact of dual physiological-aware alignment.} First, Table~\ref{tab:model_components} analyzes the contribution of our dual physiological-aware alignment strategy in \SysName. The full model, which combines Gramian-based contrastive alignment and soft lead consistency alignment, achieves the strongest overall performance. Training with only an image-text contrastive objective leads to performance degradation of over 2\% across linear probing and zero-shot classification tasks. Similarly, excluding the soft-lead consistency loss results in a decrease of approximately 2\% in linear probing, while zero-shot results are less affected, yet this still highlights the importance of enforcing inter-lead consistency during pretraining. These findings demonstrate that physiological-aware reconstruction and multimodal alignment are complementary, as reconstruction encourages preservation of fine-grained temporal morphology, while latent alignment ensures semantic consistency across modalities. We also report a baseline performance when the model is training with reconstruction only (first row) using the image encoder and signal decoder under single MSE loss, which results in a noticeable drop of 9\% in the linear probing experiments. Finally, in an additional experiment where we use only contrastive learning across the three modalities, performance decreases to $73.96 \pm 7.22$ and $69.97 \pm 8.87$ in the zero-shot and linear probing experiments, respectively, suggesting that the Gramian-based alignment better captures higher-order relationships among modalities beyond pairwise similarity.

\textbf{Impact of different modality encoders.} Table~\ref{tab:encoders} shows the performance when replacing the default text, image, and signal encoders with common alternative backbones. While performance varies slightly across choices, \SysName remains generally robust to these changes. Specifically, we observe moderately better performance with the proposed Bio-Med-CPT text encoder~\cite{jin2023medcpt} compared to Bio-ClinicalBERT~\cite{alsentzer2019publicly}, which is consistent with prior findings in MERL~\cite{liu2024zero}. For the signal encoder, using MELP results in an average performance drop of around 2.5\%, likely due to its more compact embedding dimension (e.g., 256), which may be less compatible with our large-scale multimodal alignment and decoding objectives. For the image encoder, CLIP slightly outperforms MedSigLIP, despite the latter being trained on medical imaging data (e.g., X-rays, ophthalmology, or CT/MRI) but not specifically on ECG data. Overall, these results further confirm the effectiveness of our chosen encoders, while also demonstrating that our training framework remains flexible across different encoder choices.

\begin{figure}[t]
    \centering

    \begin{subfigure}{0.11\textwidth}
        \centering
        \includegraphics[width=0.95\linewidth]{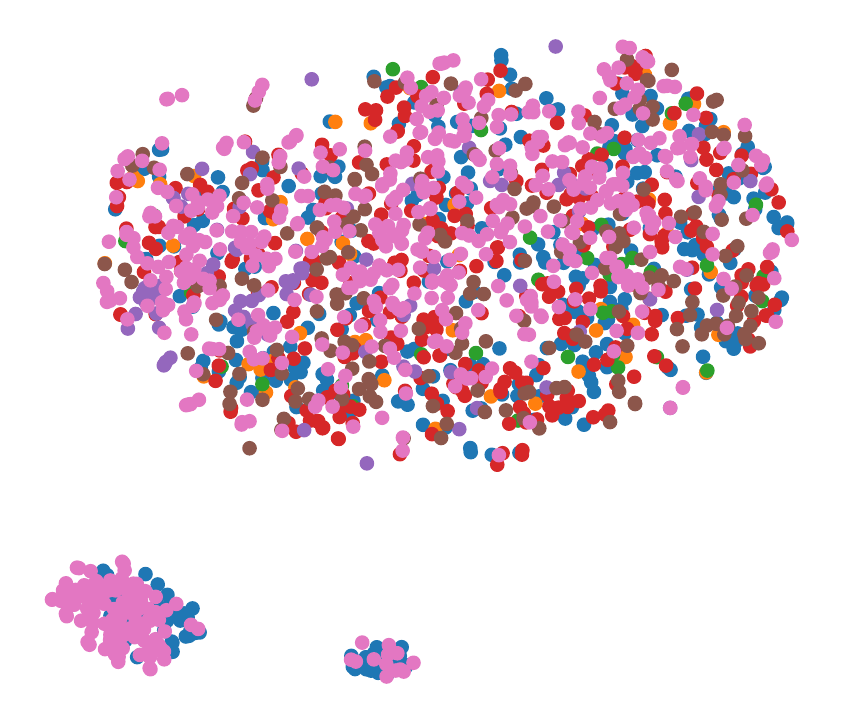}
        \caption{\scriptsize ST-MEM}
    \end{subfigure}
    \begin{subfigure}{0.11\textwidth}
        \centering
        \includegraphics[width=0.95\linewidth]{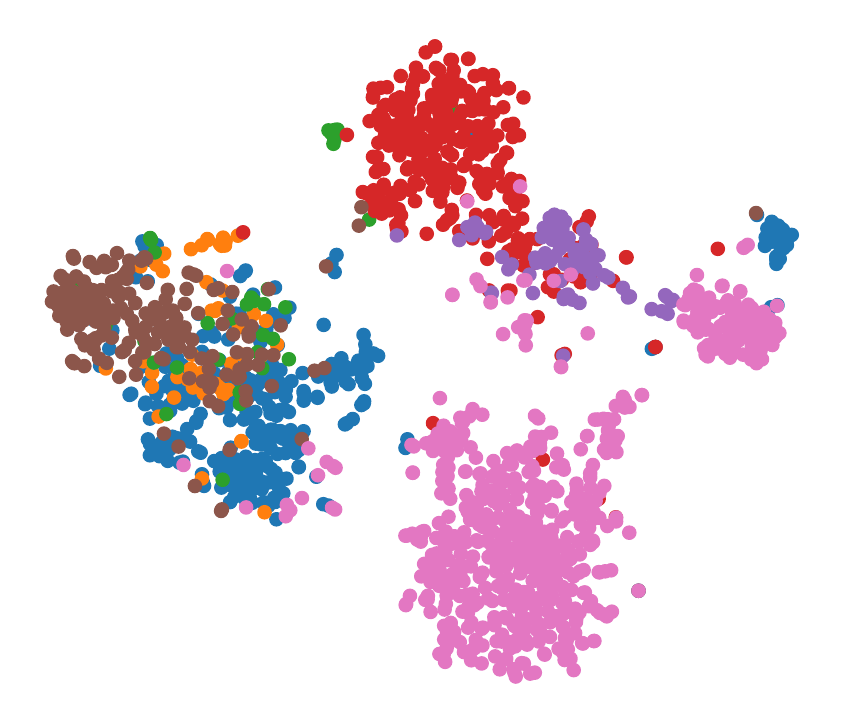}
        \caption{\scriptsize MERL}
    \end{subfigure}
    \begin{subfigure}{0.11\textwidth}
        \centering
        \includegraphics[width=0.95\linewidth]{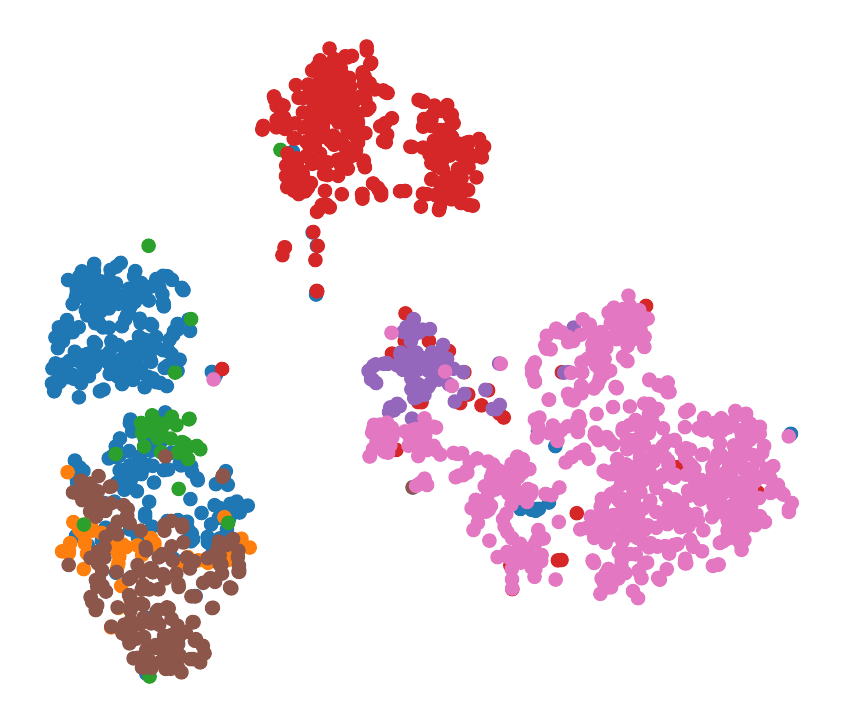}
        \caption{\scriptsize MELP}
    \end{subfigure}
    \begin{subfigure}{0.11\textwidth}
        \centering
        \includegraphics[width=0.95\linewidth]{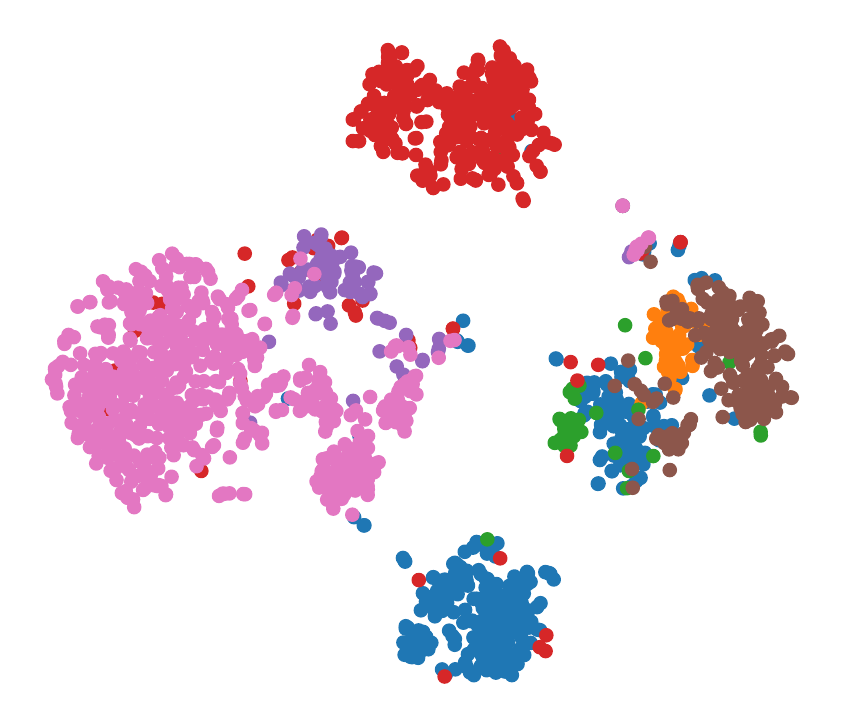}
        \caption{\scriptsize ECGFounder}
    \end{subfigure}

    \vspace{0.1cm}

    \begin{subfigure}{0.11\textwidth}
        \centering
        \includegraphics[width=0.95\linewidth]{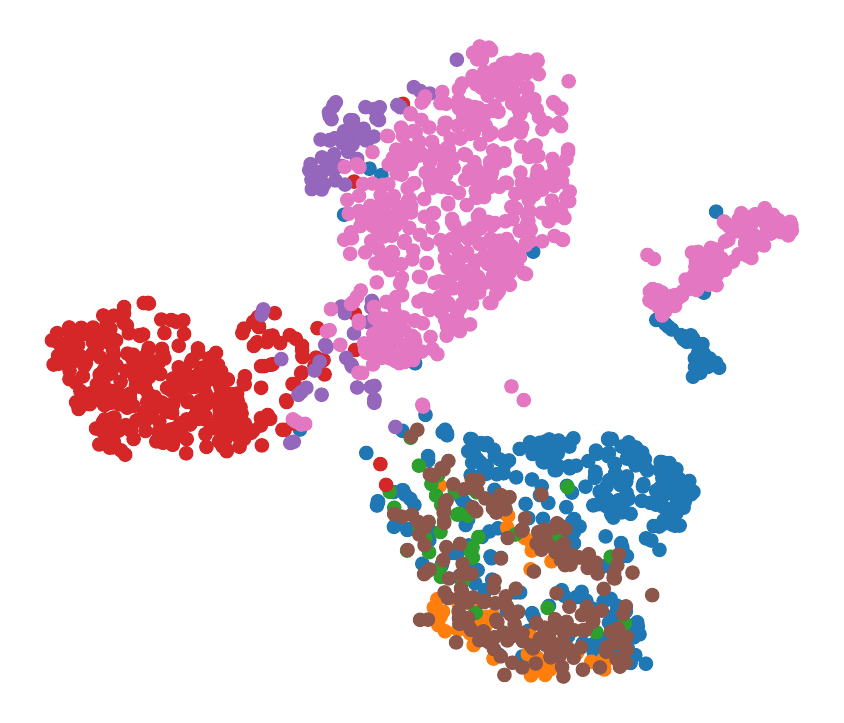}
        \caption{\scriptsize ECG-FM}
    \end{subfigure}
    \begin{subfigure}{0.11\textwidth}
        \centering
        \includegraphics[width=0.95\linewidth]{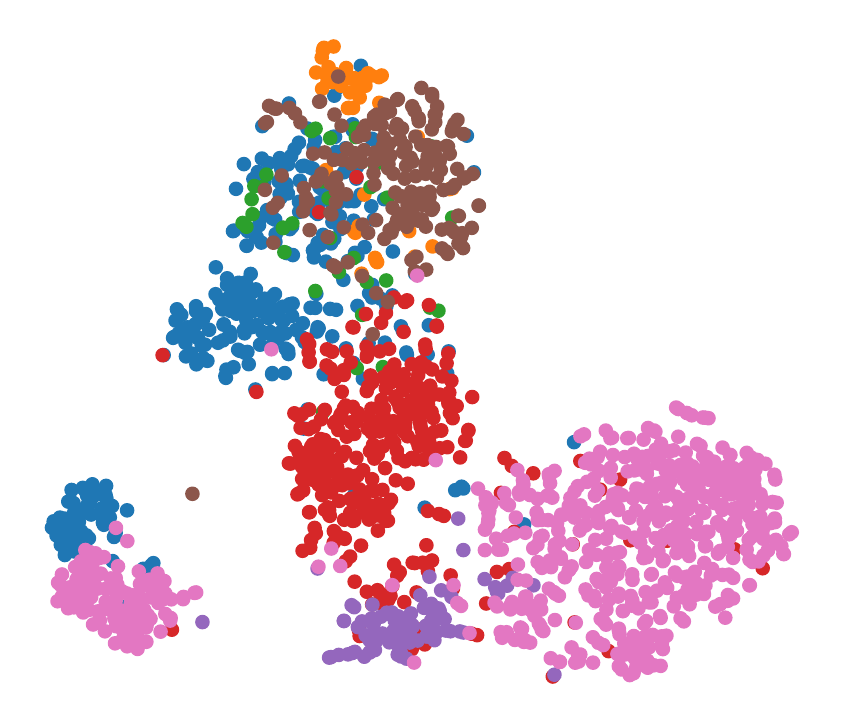}
        \caption{\scriptsize AnyECG-chat}
    \end{subfigure}
    \begin{subfigure}{0.11\textwidth}
        \centering
        \includegraphics[width=0.95\linewidth]{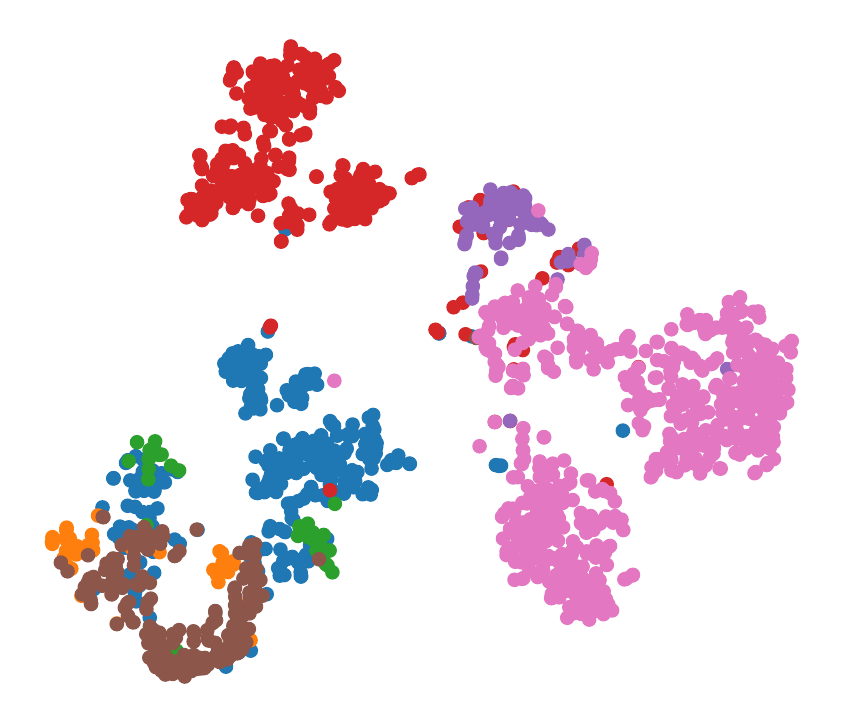}
        \caption{\scriptsize D-BETA}
    \end{subfigure}
    \begin{subfigure}{0.11\textwidth}
        \centering
        \includegraphics[width=0.95\linewidth]{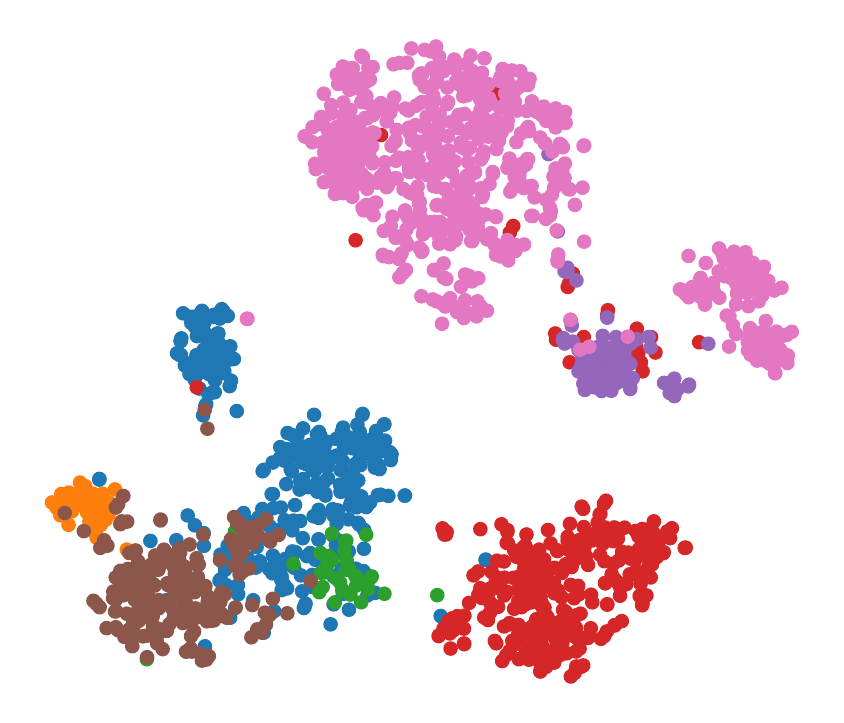}
        \caption{\scriptsize \SysName}
    \end{subfigure}

    \caption{T-SNE visualizations of representations learned by different ECG encoders on the CSN testing set. Here, \SysName deals with image input, while the others use 10s ECG signals. Each color represents a cardiac diagnosis category.}  

    \label{fig:tsne_comparison}
\end{figure}

\textbf{T-SNE visualizations.} Besides the quantitative results, we also conduct a T-SNE visualization of the learned representations on the CSN test set, which contains the selected seven cardiac conditions following \cite{liu2024zero}. As shown in Figure~\ref {fig:tsne_comparison}, compared to prior signal-reconstruction-based methods (e.g., ST-MEM), \SysName exhibits more compact intra-class clusters and clearer inter-class separation. Moreover, despite operating on image inputs, the structure of the learned embedding space closely resembles that of the state-of-the-art signal-based encoders such as ECG-FM and MELP, indicating preservation of physiologically related information.

%% file: tables/linear_probing.tex
\begin{table*}[t]
    \centering
    \setlength\tabcolsep{2.4pt}
    \caption{Linear probing performance (AUC in \%) across multiple models, datasets, and training sizes. Here, we compare the common signal foundation models as upper bounds, with different \colorbox{gray!10}{image baselines.}}

    \label{tab:linear}
    \resizebox{0.995\textwidth}{!}{
    \begin{tabular}{ll
        S[table-format=2.2]S[table-format=2.2]S[table-format=2.2]|
        S[table-format=2.2]S[table-format=2.2]S[table-format=2.2]|
        S[table-format=2.2]S[table-format=2.2]S[table-format=2.2]|
        S[table-format=2.2]S[table-format=2.2]S[table-format=2.2]|
        S[table-format=2.2]S[table-format=2.2]S[table-format=2.2]|
        S[table-format=2.2]S[table-format=2.2]S[table-format=2.2]|
        S[table-format=2.2]S[table-format=2.2]S[table-format=2.2]
    }
        \toprule
        & \multirow{2}{*}{\textbf{Methods}} & \multicolumn{3}{c}{\textbf{PTBXL-Super}}
        & \multicolumn{3}{c}{\textbf{PTBXL-Sub}}
        & \multicolumn{3}{c}{\textbf{PTBXL-Form}}
        & \multicolumn{3}{c}{\textbf{PTBXL-Rhythm}}
        & \multicolumn{3}{c}{\textbf{CPSC2018}}
        & \multicolumn{3}{c}{\textbf{CSN}} & \multicolumn{3}{c}{\textbf{Average}}\\
        \cmidrule{3-5} \cmidrule(lr){6-8} \cmidrule(lr){9-11} \cmidrule(lr){12-14} \cmidrule(lr){15-17} \cmidrule(lr){18-20} \cmidrule(lr){21-23}

        & & \textbf{1\%} & \textbf{10\%} & \textbf{100\%}
        & \textbf{1\%} & \textbf{10\%} & \textbf{100\%}
        & \textbf{1\%} & \textbf{10\%} & \textbf{100\%}
        & \textbf{1\%} & \textbf{10\%} & \textbf{100\%}
        & \textbf{1\%} & \textbf{10\%} & \textbf{100\%}
        & \textbf{1\%} & \textbf{10\%} & \textbf{100\%} & \textbf{1\%} & \textbf{10\%} & \textbf{100\%} \\
        \midrule


        \multicolumn{1}{c|}{\multirow{18}{*}{\rotatebox{90}{10s Signal Input}}} & SimCLR~\citep{chen2020simple} & 63.41 & 69.77 & 73.53 & 60.84 & 68.27 & 73.39 & 54.98 & 56.97 & 62.52 & 51.41 & 69.44 & 77.73 & 59.78 & 68.52 & 76.54 & 59.02 & 67.26 & 73.20 & 58.24 & 66.70 & 72.82 \\
        \multicolumn{1}{c|}{} & BYOL~\citep{grill2020bootstrap}  & 71.70 & 73.83 & 76.45 & 57.16 & 67.44 & 71.64 & 48.73 & 61.63 & 70.82 & 41.99 & 74.40 & 77.17 & 60.88 & 74.42 & 78.75 & 54.20 & 71.92 & 74.69  & 55.78 & 70.61 & 74.92\\
        \multicolumn{1}{c|}{} & BarlowTwins~\citep{zbontar2021barlow} & 72.87 & 75.96 & 78.41 & 62.57 & 70.84 & 74.34 & 52.12 & 60.39 & 66.14 & 50.12 & 73.54 & 77.62 & 55.12  & 72.75 & 78.39 & 60.72 & 71.64 & 77.43 & 58.92 & 70.85 & 75.39 \\
        \multicolumn{1}{c|}{} & MoCo-v3~\citep{chen2021empirical} & 73.19 & 76.65 & 78.26 & 55.88 & 69.21 & 76.69 & 50.32 & 63.71 & 71.31 & 51.38 & 71.66 & 74.33 & 62.13 & 76.74 & 75.29 & 54.61 & 74.26 & 77.68 & 57.92 & 72.04 & 75.59 \\
        \multicolumn{1}{c|}{} & SimSiam~\citep{chen2021exploring} & 73.15 & 72.70 & 75.63 & 62.52 & 69.31 & 76.38 & 55.16 & 62.91 & 71.31 & 49.30 & 69.47 & 75.92 & 58.35 & 72.89 & 75.31 & 58.25 & 68.61 & 77.41 & 59.46 & 69.32 & 75.33\\
        \multicolumn{1}{c|}{} & TS-TCC~\citep{eldele2021time} & 70.73 & 75.88 & 78.91 & 53.54 & 66.98 & 77.87 & 48.04 & 61.79 & 71.18 & 43.34 & 69.48 & 78.23 & 57.07 & 73.62 & 78.72 & 55.26 & 68.48 & 76.79& 54.66 & 69.37 & 76.95\\
        \multicolumn{1}{c|}{} & CLOCS~\citep{kiyasseh2021clocs} & 68.94 & 73.36 & 76.31 & 57.94 & 72.55 & 76.24 & 51.97 & 57.79 & 72.65 & 47.19 & 71.88 & 76.31 & 59.59 & 77.78 & 77.49 & 54.38 & 71.93 & 76.13 & 56.67 & 70.88 & 75.86\\
        \multicolumn{1}{c|}{} & ASTCL~\citep{wang2023adversarial} & 72.51 & 77.31 & 81.02 & 61.86 & 68.77 & 76.51 & 44.14 & 60.93 & 66.99 & 52.38 & 71.98 & 76.05 & 57.90  & 77.01 & 79.51 & 56.40 & 70.87 & 75.79& 57.53 & 71.14 & 75.98\\
        \multicolumn{1}{c|}{} & CRT~\citep{zhang2023self} & 69.68 & 78.24 & 77.24 & 61.98 & 70.82 & 78.67 & 46.41 & 59.49 & 68.73 & 47.44 & 73.52 & 74.41 & 58.01 & 76.43 & 82.03 & 56.21 & 73.70 & 78.80 & 56.62 & 72.03 & 76.65\\
        \multicolumn{1}{c|}{} & ST-MEM~\citep{na2024guiding} & 61.12 & 66.87 & 71.36 & 54.12 & 57.86 & 63.59 & 55.71 & 59.99 & 66.07 & 51.12 & 65.44 & 74.85 & 56.69 & 63.32 & 70.39 & 59.77 & 66.87 & 71.36& 56.42 & 63.39 & 69.60\\
        \multicolumn{1}{c|}{} & MERL~\citep{liu2024zero} & 82.39 & 86.27 & 88.67 & 64.90 & 80.56 & 84.72 & 58.26 & 72.43 & 79.65 & 53.33 & 82.88 & 88.34 & 70.33 & 85.32 & 90.57 & 66.60 & 82.74 & 87.95& 65.97 & 81.70 & 86.65\\
        \multicolumn{1}{c|}{} & ESI~\citep{esi} & 62.85 & 78.07 & 83.22 & 63.78 & 71.45 & 78.54 & 60.76 & 64.19 & 74.19 & 60.93 & 70.56 & 78.48 & 69.12 & 77.50 & 83.03 & 55.29 & 68.41 & 74.42 & 62.12 & 71.70 & 78.65\\
        \multicolumn{1}{c|}{} & Heartlang~\citep{heartlang} & 73.06 & 84.20 & 87.96 & 65.50 & 77.91 & 84.51 & 59.08 & 68.86 & 81.25 & 53.99 & 80.57 & 91.32 & 65.97 & 80.26 & 88.01 & 57.64 & 68.71 & 76.34 & 62.54 & 76.75 & 84.90\\
        \multicolumn{1}{c|}{} & ECG-FM~\citep{ecgfm} & 71.92 & 82.17 & 85.94 & 65.65 & 77.51 & 83.94 & 58.76 & 68.90 & 78.84 & 76.71 & 89.14 & 95.13 & 72.68 & 88.53 & 92.92 & 67.34 & 84.64 & 92.32& 68.84 & 81.82 & 88.18\\
        \multicolumn{1}{c|}{} & AnyECG-chat~\cite{anychat} & 79.20 & 84.74 & 86.68 & 74.28 & 79.14 & 84.04 & 64.84 & 74.69 & 79.61 & 80.66 & 92.37 & 96.00 &  80.03 & 89.95 & 92.79 & 75.25 & 87.10 & 90.89 & 75.71 & 84.66 & 88.34\\
        \multicolumn{1}{c|}{} & ECGFounder~\citep{ecgfounder} & 85.11 & 88.68 & 90.74 & 80.72 & 84.16 & 87.85 & 72.18 & 81.82 & 86.44 & 85.45 & 94.28 & 97.52 & 67.90 & 80.63 & 89.84 & 70.43 & 86.66 & 93.42 & 76.96 & 86.04 & 90.97\\
        \multicolumn{1}{c|}{} & MELP~\citep{melp} & 82.83 & 88.81 & 89.97 & 76.46 & 85.27 & 87.93 & 68.58 & 80.76 & 85.21 & 81.89 & 91.87 & 96.78 & 84.91 & 94.29 & 95.83 & 80.69 & 90.55 & 93.49& 79.23 & 88.59 & 91.53\\
        \multicolumn{1}{c|}{} & D-BETA~\citep{dbeta} & 84.09 & 88.86 & 89.84 & 77.36 & 81.20 & 86.74 & 72.43 & 79.56 & 84.60 & 86.56 & 93.94 &  97.23& 91.34 & 94.83 & 96.51 & 81.28 & 91.26 & 94.43& 82.18 & 88.28 & 91.56\\

        \midrule


        \rowcolor{shadegray}

          \multicolumn{1}{c|}{\multirow{6}{*}{\rotatebox{90}{}}}  &  CLIP~\cite{clip}& 70.46 & 80.22 & 83.65 & 64.36 & 67.87 & 78.79 & 48.25 & 54.50 & 73.38 & 60.06 & 71.93 & 81.33 & 61.08 & 75.59 & 87.33 & 53.31 & 62.28 & 84.10& 59.59 & 68.73 & 81.43
 \\

        \rowcolor{shadegray}

        \multicolumn{1}{l|}{}& MedSigLIP~\citep{medsiglip} & 73.24 & 78.41 & 81.53 & 65.25 & 65.76 & 74.56 & 46.76 & 57.90 & 70.85 & 67.37 & 71.74 & 80.11 & 59.94 & 76.03 & 86.00 & 49.52 & 62.82 & 79.04 & 60.35 &68.78 & 78.68
\\

        \rowcolor{shadegray}

        \multicolumn{1}{l|}{}& DIP + D-BETA & 70.27 & 77.86 & 80.70 & 67.46 & 71.42 & 78.03 & 56.67 & 65.14 & 70.00 & 65.03 & 81.92 & 85.28 & 62.22 & 69.32 & 73.62 & 63.27 & 71.60 & 82.92 & 64.15 & 72.88 & 78.43
\\
        \rowcolor{shadegray}

        \multicolumn{1}{l|}{}& nnUNet + D-BETA ~\citep{nnunet} & 76.08 & 82.89 & 84.66 & 71.48 & 74.63 & 81.50 & 56.22 & 70.03 & 78.09 & 76.19 & 84.62 & 93.07 & 73.30 & 84.72 & 91.49 & 69.68 & 78.48 & 90.13 & 70.49  & 79.23 & 86.49
\\

        \rowcolor{shadegray}
        \multicolumn{1}{l|}{}& Open-Digitizer + D-BETA~\cite{opendigitizer} & 79.24 & 85.32 & 86.30 & 71.72 & 77.85 & 83.47 & 56.96 & 71.37 & 77.02 & 77.77 & 87.21 & 90.36 & 74.97 & 89.36 & 92.74 & 69.16 & 77.78 & 89.47 &71.64&81.48&86.56 \\
        \midrule
        \rowcolor{shadegray}
        
        \multicolumn{1}{l|}{}& \textbf{\SysName} & 82.67 & 88.40 & 90.86 & 69.02 & 79.42 & 85.34 & 57.40 & 77.41 & 84.08 & 74.74 & 85.01 & 93.31 & 77.71 & 88.83 & 94.64 & 70.04 & 87.96 & 94.20 & 71.93 & 84.51& 90.41\\

        \bottomrule
    \end{tabular}
    }
\end{table*}

%% file: tables/zero-shot.tex
\begin{table*}[t]
    \centering
    \setlength\tabcolsep{3.8pt}
    \scriptsize
    
    \caption{Zero-shot performance (AUC in \%) across multiple models and datasets. Here, we reported three signal foundation models: MERL, D-BETA, and MELP as upper bounds (signal-text), comparing with different \colorbox{gray!10}{image baselines} (image-text).}
    
    \label{tab:zerozhot1}
    \begin{tabular}{lcccccccc}
    
        \toprule
        
        \textbf{Methods} & \textbf{ECG Input} &
        \textbf{PTBXL-Super} & \textbf{PTBXL-Sub} & \textbf{PTBXL-Form} &
        \textbf{PTBXL-Rhythm} & \textbf{CPSC2018} & \textbf{CSN} &
        \textbf{Average} \\
        
        \midrule

        MERL~\cite{liu2024zero}   & 10s Signal & 74.2 & 75.7 & 65.9 & 78.5 & 82.8 & 74.4 & 75.3 \\     
        D-BETA~\cite{dbeta}       & 10s Signal & 76.2 & 75.9 & 66.1 & 88.6 & 80.1 & 76.3 & 77.1  \\
        MELP~\cite{melp}          & 10s Signal & 76.2 & 81.2 & 69.1 & 85.4 & 84.2 & 77.6 & 79.0 \\
        \midrule

        \rowcolor{shadegray}
        DIP + D-BETA             & Image  & 61.2 & 63.4 & 53.1 & 76.5 & 58.4 & 66.3 & 63.2 \\
        \rowcolor{shadegray}
        nnU-Net + D-BETA \cite{nnunet}
                                 & Image  & 61.1 & 65.2 & 58.3 & 75.9 & 72.0 & 66.5 & 66.5 \\
        \rowcolor{shadegray}
        Open-Digitizer + D-BETA \cite{opendigitizer}
                                 & Image  & 67.3 & 64.8 & 58.6 & 82.7 & 73.6 & 64.4 & 68.6 \\
        \midrule

        \rowcolor{shadegray}
               \textbf{\SysName }        & Image  & 77.2 & 76.7 & 65.1 & 84.0 & 80.9 & 71.1 & 75.8\\
        \bottomrule
    \end{tabular}
\end{table*}

\begin{table*}[t]
    \centering
    \setlength\tabcolsep{2.4pt}
    \scriptsize
    \caption{ECG interpretation comparisons (AUC in \%) on the same source CODE-test dataset: Human experts\cite{ribeiro_automatic_2020} vs. different 10s signal foundation models (zero-shot signal-text) vs. different \colorbox{gray!10}{image baselines} (zero-shot image-text).}

    \label{tab:zeroshot2}
    \begin{tabular}{ccccccccc}
        \toprule
\multicolumn{3}{c}{\textbf{Human Expert's Interpretation}} &
\multicolumn{2}{c}{\textbf{Signal-based Models}} &
\multicolumn{4}{c}{\cellcolor{shadegray}\textbf{Image-based Models}} \\
\cmidrule(lr){1-3} \cmidrule(lr){4-5} \cmidrule(lr){6-9}
        Cardio Resident & Emergency Resident & Medical Student & MERL & D-BETA & \cellcolor{shadegray}DIP + D-BETA&  \cellcolor{shadegray}nnUnet + D-BETA &  \cellcolor{shadegray}Open-Digitizer + D-BETA  & \cellcolor{shadegray}\textbf{\SysName} \\
        \midrule
        92.07 & 90.52 & 93.61 & 85.14 & 96.79
& \cellcolor{shadegray}  64.97 
& \cellcolor{shadegray} 85.20
& \cellcolor{shadegray} 94.04 
& \cellcolor{shadegray}94.78 \\

        \bottomrule
    \end{tabular}
\end{table*}

%% file: tables/domain_shift.tex
\begin{table*}[h]
    \centering
    \caption{
    Performance under data distribution shift. Here, \SysName deals with image input, while the others use 10s signals.
    }
    \label{tab:domainshift}
    \resizebox{0.92\textwidth}{!}{
        \begin{tabular}{l c c c c c c c c c}
        \toprule
        Source Domain & \multirow{2}{*}{Zero-shot} & \multirow{2}{*}{Training Ratio} 
        & \multicolumn{2}{c}{PTBXL-Super}
        & \multicolumn{2}{c}{CPSC2018}
        & \multicolumn{2}{c}{CSN} 
        & \multirow{2}{*}{Average} \\
        Target Domain & ~ & ~ 
        & CPSC2018 & CSN & PTBXL-Super & CSN & PTBXL-Super & CPSC2018 \\
        \midrule
        SimCLR~\cite{chen2020simple}      
        &\xmark & 100\% & 69.62 & 73.05 & 56.65 & 66.36 & 59.74 & 62.11 & 65.22 \\
        BYOL~\cite{grill2020bootstrap}      
        &\xmark & 100\% & 70.27 & 74.01 & 57.32 & 67.56 & 60.39 & 63.24 & 65.63 \\
        BarlowTwins~\cite{zbontar2021barlow} 
        &\xmark & 100\% & 68.98 & 72.85 & 55.97 & 65.89 & 58.76 & 61.35 & 64.13 \\
        MoCo-v3~\cite{chen2021empirical}     
        &\xmark & 100\% & 69.41 & 73.29 & 56.54 & 66.12 & 59.82 & 62.07 & 64.21 \\
        SimSiam~\cite{chen2021exploring}     
        &\xmark & 100\% & 70.06 & 73.92 & 57.21 & 67.48 & 60.23 & 63.09 & 65.33 \\
        TS-TCC~\cite{eldele2021time}      
        &\xmark & 100\% & 71.32 & 75.16 & 58.47 & 68.34 & 61.55 & 64.48 & 66.55 \\
        CLOCS~\cite{kiyasseh2021clocs}       
        &\xmark & 100\% & 68.79 & 72.64 & 55.86 & 65.73 & 58.69 & 61.27 & 63.83 \\
        ASTCL~\cite{wang2023adversarial}       
        &\xmark & 100\% & 69.23 & 73.18 & 56.61 & 66.27 & 59.74 & 62.12 & 64.19 \\
        CRT~\cite{zhang2023self}         
        &\xmark & 100\% & 70.15 & 74.08 & 57.39 & 67.62 & 60.48 & 63.33 & 65.51 \\
        ST-MEM~\cite{na2024guiding}  &\xmark & 100\% & 76.12 & 84.50 & 62.27 & 75.19 & 73.05 & 64.66 & 72.63 \\
        D-BETA~\citep{dbeta}  & \checkmark & 0\% & 72.09 & 79.11 & 77.12 & 82.91 & 76.24 & 80.10 & 77.93 \\
        MERL~\cite{liu2024zero}  & \checkmark & 0\%   & 88.21 & 78.01 & 76.77 & 76.56 & 74.15 & 82.86 & 79.42 \\
        MELP~\cite{melp} & \checkmark &  0\%  & 87.75 & 74.11 & 77.89 & 80.32 & 74.67 & 82.72 & 79.58 \\

        \midrule
        \rowcolor{shadegray}
                \textbf{\SysName} & \checkmark &  0\% & 84.27 & 73.26 & 84.37 & 80.31 & 77.22 & 80.88 & 80.05 \\
        \bottomrule
        \end{tabular}
    }
\end{table*}

%% file: tables/ablations.tex
\begin{table}[t]

\setlength{\tabcolsep}{10pt}
\footnotesize
\centering
\caption{Effects of dual physiological-aware alignment.}
\label{tab:model_components}
\begin{tabular}{@{}ccccccccccc@{}}
\toprule
$\mathcal{L}_{\text{ctr}}$ & $\mathcal{L}_{\text{gram}} $ &  $\mathcal{L}_{\text{rule}}$  & \textbf{Zero-shot} & \textbf{Linear probing} \\ 
\midrule

&  &  & --  & $62.71\pm9.62$ \\

\checkmark  & &  & $73.62 \pm 7.25$  & $68.35 \pm 8.19$  \\

\checkmark  & \checkmark  & & $75.11 \pm 6.70 $& $69.76 \pm 8.94$ \\

\checkmark  & \checkmark  & \checkmark &  $75.83 \pm 6.79$ & $71.92 \pm 8.72$  \\

\bottomrule
\end{tabular}
\end{table}

\begin{table}[ht!]
\setlength{\tabcolsep}{6pt}
\footnotesize
\centering
\caption{Effects of different encoders across modalities.}
\label{tab:encoders}
\begin{tabular}{@{}llcc@{}}
\toprule
\textbf{Modality} & \textbf{Replaced Encoder} & \textbf{Zero-shot} & \textbf{Linear probing} \\
\midrule

Text
& Bio-ClinicalBERT~\cite{alsentzer2019publicly} & $74.38 \pm 8.15$ & $70.95 \pm 8.82$ \\

Image
& MedSigLIP~\cite{medsiglip} & $73.52 \pm 7.86$ & $70.35 \pm 8.81$ \\

Signal & MELP~\cite{melp}   & $73.58 \pm 8.77$ & $69.47 \pm 8.99$ \\
\midrule
 \textbf{\SysName}&& $75.83 \pm 6.79$ & $71.92 \pm 8.72$ \\

\bottomrule
\end{tabular}
\end{table}

%% file: sections/Conclusion.tex
\section{Conclusion}

We presented a multimodal framework for learning ECG image representations. By leveraging two-level domain-informed alignments of image and signal-text modalities, our method learns physiologically grounded features without manual annotations. Extensive evaluations across diverse downstream tasks show that our image representations achieve performance comparable to strong existing foundation models. These results highlight the potential of pretraining ECG images to develop a supportive tool in cardiovascular diagnosis. Future work will leverage our pretraining framework to further scale on emerging ECG data and enable validation under real imaging conditions as such datasets become accessible.

%% file: sections/Appendix.tex
\newpage
\appendix
\twocolumn
\label{appendix}

\begin{figure*}[h]
    \centering

    \begin{subfigure}[t]{\textwidth}
        \centering
        \includegraphics[width=0.9\linewidth]{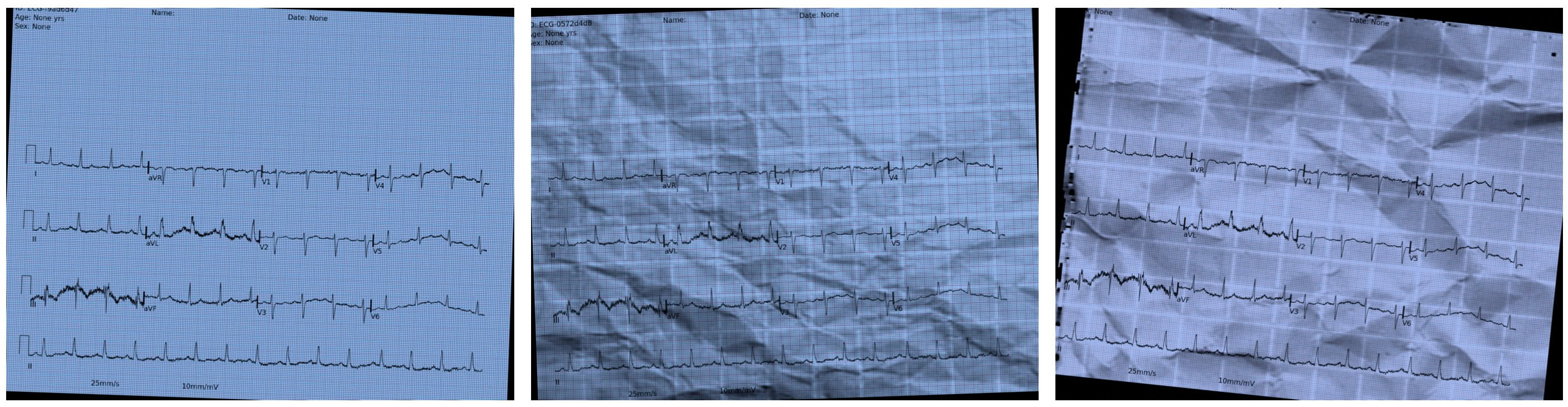}
        \label{fig:data1}
    \end{subfigure}

    \vspace{0.5em}

    \begin{subfigure}[t]{\textwidth}
        \centering
        \includegraphics[width=0.9\linewidth]{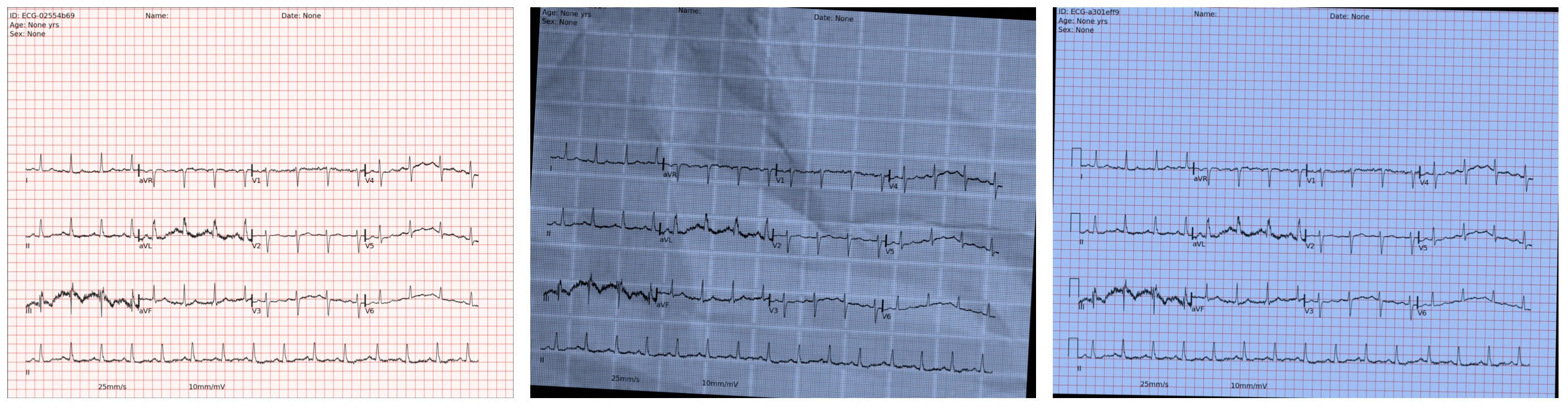}
        \label{fig:data2}
    \end{subfigure}

    \caption{Examples of image augmentations during pretraining.}
    \label{fig:data}
\end{figure*}

\newcommand{\toptitlebar}{
  \hrule height 0.5pt
  \vskip 0.1in
}

\newcommand{\bottomtitlebar}{
  \vskip 0.1in
  \hrule height 0.5pt
  \vskip 0.1in
}

\section{Additional Model Details}

\subsection{Modality Encoders and Signal Decoder}

Our framework leverages three modality-specific encoders, each producing fixed-dimensional representations that are subsequently aligned in a shared embedding space.

\textbf{Image Encoder.} We initialize the image encoder from the CLIP vision encoder~\cite{clip}, a pretrained model that provides strong visual representations for general images. To efficiently adapt the encoder to the ECG image domain while preserving pretrained knowledge, we employ Low-Rank Adaptation (LoRA)~\cite{hu2022lora}. Then, given an ECG image $\mathcal{X}_{\text{img}}$, the encoder $f_{\theta}$ produces a ECG image representation:
\begin{equation}
    \mathbf{z}_{\text{img}} = f_{\theta}(\mathcal{X}_{\text{img}}) \in \mathbb{R}^{d_{\text{img}}}
\end{equation}
where $d_{\text{img}} = 1024$. Two separate linear projectors with Tanh activation then map this representation to the shared signal embedding space: $\text{Proj}_{\text{rec}}$ for signal reconstruction and $\text{Proj}_{\text{ctr}}$ for contrastive alignment, both outputting $d_{\text{sig}} = 768$ dimensions. It is worth noting that our trained \SysName's image encoder clearly surpassed the original CLIP (see our results in the main text), which is used in PULSE~\cite{pulse} and GEM~\cite{gem} for their image encoder in textual generation tasks. 

\textbf{Signal Encoder.} We employ D-BETA~\cite{dbeta} as the signal encoder, a recent ECG foundation model pretrained on large-scale 12-lead ECG data. This encoder is shown to be robust across different datasets and tasks, as can be seen in our results sections. Throughout our pretraining, it remains \textit{frozen}  and serves as a teacher model, providing high-quality signal representations that guide the image encoder learning. Given a golden standard 12-lead ECG signal $\mathcal{X}_{\text{sig}} \in \mathbb{R}^{12 \times T}$, the encoder produces ECG signal representation:
\begin{equation}
    \mathbf{z}_{\text{sig}} = g_{\phi}(\mathcal{X}_{\text{sig}}) \in \mathbb{R}^{d_{\text{sig}}}
\end{equation}
where $d_{\text{sig}} = 768$. By distilling knowledge from this frozen encoder, we enable the image encoder to learn representations compatible with signal-based models without requiring paired labeled data.

\textbf{Text Encoder.} For encoding clinical text reports, we use Bio-Med-CPT~\cite{jin2023medcpt}, a popularly used domain-specific medical language model in ECG works~\cite{liu2024zero, melp}. Here, the text encoder is also frozen during training, as suggested in METS~\cite{li2024frozen}. Given a text report $\mathcal{X}_{\text{txt}}$, we extract the representation and project it to the shared space:
\begin{equation}
    \mathbf{z}_{\text{txt}} = \text{Proj}_{\text{txt}}(h_{\psi}(\mathcal{X}_{\text{txt}})) \in \mathbb{R}^{d_{\text{sig}}}
\end{equation}
where $\text{Proj}_{\text{txt}}$ is a linear projector with Tanh activation mapping from $d_{\text{txt}}$ to $d_{\text{sig}}$.

\textbf{Signal Decoder.} Next, we encourage the image encoder to capture fine-grained physiological structure by introducing a signal decoder that recovers the underlying gold-standard 12-lead ECG signals. Here, signal reconstruction is adopted because it explicitly enforces preservation of temporal morphology, which is central to clinical ECG interpretation yet difficult to recover from generic visual features or short per-lead temporal contexts commonly present in ECG images. We formulate reconstruction as a sequence generation problem and implement a Transformer-based decoder to model long-range temporal dependencies. Concretely, the image representation $\mathbf{z}_{\text{img}}$ is first projected into a reconstruction latent space, yielding $\mathbf{z}_{\text{rec}} = \text{Proj}_{\text{rec}}(\mathbf{z}_{\text{img}})$, which serves as global conditioning for signal generation. We initialize a set of $N_p = \lceil T / P \rceil$ learnable query tokens, where $T$ denotes the target signal length (i.e., 5000) and $P$ the patch size at 8. The projected latent is added to each query token, while learnable positional embeddings encode temporal order. These tokens are then processed by a Transformer encoder with $L=12$ layers, hidden dimension $d=768$, and $H=12$ attention heads, enabling joint modeling of temporal structure and inter-lead correlations. A final linear projection maps each token to $C \times P$ values, which are reshaped and concatenated to form the reconstructed ECG signal $\hat{\mathcal{X}}_{\text{sig}} \in \mathbb{R}^{12 \times T}$. During training, we apply random masking to a subset of query tokens, replacing them with a learnable mask embedding. This strategy prevents the decoder from relying on fixed positional cues and encourages reconstruction to be driven primarily by the global image-derived representation, thereby improving robustness and generalization.

\subsection{Gramian-based ECG-Text Contrastive Learning}

As described in Section Method, we incorporate a Gramian-based alignment as an auxiliary objective to regularize our multimodal representation learning. Unlike prior formulations that enforce strict pairwise similarity constraints, we reinterpret the Gramian as a signal-text distillation mechanism that transfers higher-order relational structure from physiologically grounded ECG signal embeddings to image-aligned representations. Concretely, the Gramian captures global covariance patterns within modalities, encoding semantic dependencies that reflect underlying cardiac information from additional clinical texts and gold standard signals rather than instance-level correspondence. In our framework, this serves as a physiological prior that complements contrastive image-text alignment: while contrastive learning emphasizes discriminative instance separation (strongly supporting zero-shot image-text experiments), the Gramian constraint preserves intrinsic signal geometry. Besides the ablation studies, we conduct an additional zero-shot experiment on the CODE-test dataset by selectively removing either the image-text contrastive loss or the Gramian-based loss from the best setting. The full model achieves the best zero-shot performance at 94.8\% AUC, while removing contrastive learning or Gramian alignment results in clear degradation to  85.2\% and 90.3\%, respectively.

\section{Additional Training Details.}
\label{appendix:data}

\subsection{Dataset Splits}

Table~\ref{tab:training_config} summarizes dataset statistics and training configurations used throughout pretraining and downstream evaluation. For pretraining, after preprocessing the ECG signals and normalizing the clinical notes from the dataset~\cite{gow2023mimic}, we split it into training and validation sets using a 9:1 ratio, resulting in 710,560 training samples and 78,951 validation samples. For downstream benchmarks, we adopt the official or commonly used train/validation/test splits for each dataset to ensure fair comparability with prior work~\cite{liu2024zero}. In particular, CODE-test\cite{ribeiro_automatic_2020} is used exclusively for zero-shot evaluation and therefore contains no training or validation split.

\subsection{ECG Image Synthesis}
\label{appendix:augmentation}

To the best of our knowledge, large-scale ECG image-text-signal datasets remain scarce, which limits direct support for multimodal training goals. Therefore, we customize a commonly used pretraining ECG signal-text dataset (i.e., MIMIC IV ECG~\cite{gow2023mimic}), where ECG images are rendered from raw 12-lead signals using a configurable rendering pipeline, while keeping pairs with the clinical notes. Given a signal $\mathcal{X}_{\text{sig}} \in \mathbb{R}^{12 \times T}$, we generate a realistic ECG printout that emulates clinical ECG recordings on-the-fly during the pretraining process. Moving on, our synthesis pipeline is based on the tool~\cite{ecgkit}, which is a widely-used realistic ECG image generation pipeline~\cite{pulse}. 

We produce images using a standard clinical layout in which the six limb leads (I, II, III, aVR, aVL, aVF) and six precordial leads (V1 to V6) are arranged in a $3 \times 4$ grid, with lead~II additionally displayed as a continuous rhythm strip (at 10 seconds, other leads as 2.5 seconds). Each image is generated with a calibrated grid background (typically at a paper speed of 25 mm/s and an amplitude of 10 mm/mV), lead annotations, and optional patient metadata. To further emulate real-world acquisition and archival conditions, stochastic augmentations are applied during rendering, including geometric perturbations, noise and artifact injection, color and contrast variations, and grid style changes. This online synthesis avoids storing redundant image copies while ensuring high diversity and robustness of training samples. We provide examples of data augmentation effects in Figure~\ref{fig:data}.

Regarding the evaluation, systematic tests on large-scale real ECG image datasets remain an important direction for future work, as suitable datasets become publicly available for our work. However, we emphasize that our primary goal is to support a training framework that studies ECG image learning, rather than to strictly provide full in-the-wild evaluations on real-world ECG photographs. Our model provides a general image representation for subsequent fine-tuning on downstream tasks before a final deployable clinical utility.

\begin{figure*}[t]
    \centering
    \includegraphics[width=0.65\linewidth]{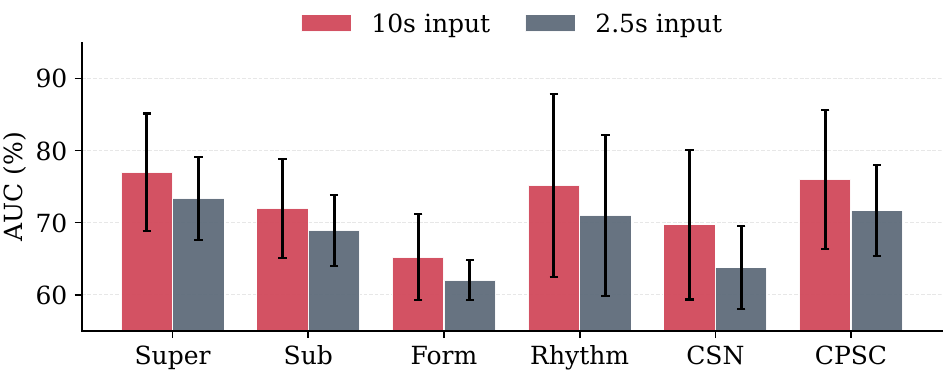}
    \caption{Linear probing performance comparing 10\,s and 2.5\,s ECG signal inputs across six datasets, averaged over seven ECG foundation models.}
    \label{fig:length}
\end{figure*}

\subsection{Linear Probing Experiments}

\begin{table*}[h]
\centering
\caption{Details on data and training configurations.}
\label{tab:training_config}
\scalebox{0.8}{
\begin{tabular}{lcccccccc}
\toprule[1.2pt]
Dataset & \# Categories & Train & Valid & Test & Optimizer & \# Epoch & Batch size & Learning rate \\ 
\midrule[1.2pt]
MIMIC-IV-ECG~\cite{gow2023mimic} & -- & 710,560 & 78,951 & -- & AdamW & -- & 80 & 0.0005 \\
\midrule
PTBXL-Super \cite{wagner2020ptb} & 5 & 17,084 & 2,146 & 2,158 & AdamW & 100 & 16 & 0.001 \\
PTBXL-Sub \cite{wagner2020ptb} & 23 & 17,084 & 2,146 & 2,158 & AdamW & 100 & 16 & 0.001 \\
PTBXL-Form \cite{wagner2020ptb} & 19 & 7,197 & 901 & 880 & AdamW & 100 & 16 & 0.001 \\
PTBXL-Rhythm \cite{wagner2020ptb} & 12 & 16,832 & 2,100 & 2,098 & AdamW & 100 & 16 & 0.001 \\
CPSC2018 \cite{liu2018open} & 9 & 4,950 & 551 & 1,376 & AdamW & 100 & 16 & 0.001 \\
CSN \cite{zheng2022large} & 38 & 16,546 & 1,860 & 4,620 & AdamW & 100 & 16 & 0.001 \\
CODE-test~\citep{ribeiro_automatic_2020} & 6 & -- & -- & 827 & -- & -- & -- & -- \\
\bottomrule[1.2pt]
\end{tabular}
}
\end{table*}
We provide additional details on the linear probing experiments across different downstream datasets in Table~\ref{tab:training_config}. Following~\cite{liu2024zero}, we freeze the pretrained encoder and train a linear classifier using 100 epochs, the AdamW optimizer with a learning rate of $1\times10^{-3}$ and a batch size of 16 for all downstream tasks. This protocol is applied consistently across all methods to ensure fair comparisons.

In addition to standard linear probing with full-length signals, we further investigate the impact of signal incompleteness by comparing downstream performance using 2.5-second and 10-second ECG signal inputs, reporting average results from seven signal models~\cite{melp, dbeta, anychat, esi, heartlang, ecgfm, ecgfounder}. As shown in Figure~\ref{fig:length}, reducing the available temporal context from 10 seconds to 2.5 seconds consistently degrades performance across all six datasets by about 5\%. \textit{This observation highlights a challenge: the downstream performance of the existing signal foundation models is closely coupled to signal length, and truncated or incomplete recordings can substantially impair representation quality.} Meanwhile, ECG images often do not explicitly encode a fixed temporal duration in the same manner. This comparison underscores an inherent robustness advantage of image-based ECG representations in real-world scenarios, and avoid usage of under-optimal methods that combine signal foundation models with image-to-signal conversions.